# Image Classification in Arabic: Exploring Direct English to Arabic Translations

**Abdulkareem Alsudais**
Information Systems Department, College of Computer Engineering and Sciences, Prince Sattam Bin Abdulaziz University, Al Kharj, Saudi Arabia
e-mail: A.Alsudais@psau.edu.sa

**ABSTRACT** Image classification is an ongoing research challenge. Most of the current research focuses on image classification in English with very little research in Arabic. Expanding image classification to Arabic has several applications and benefits. This paper investigates the accuracy of direct translations of English labels that are available in ImageNet, a database of images labeled in English that is commonly used in computer vision research, to Arabic. A dataset comprised of 2,887 labeled images was constructed by randomly selecting images from ImageNet. All of the labels were translated to Arabic using an online translation service. The accuracy of each translation was evaluated by a human judge. Results indicated that 65.6% of the generated Arabic labels were accurate with the highest results achieved when the labels consisted of only one word. This study makes three important contributions to the image classification literature: (1) it determines a baseline level of accuracy for image classification in Arabic algorithms; (2) it provides 1,910,935 images classified with accurate Arabic labels (based on accurately labeling 1,895 images that consist of 1,643 unique synsets); and (3) it measures the accuracy of translations of image labels in ImageNet to Arabic.

## I. INTRODUCTION

In recent years, advances in artificial intelligence research have been significant. One area that has seen major development is the study of computer vision, specifically image processing and classification. The objective of image classification algorithms is to generate an accurate label or a group of labels for an image that captures the content(s) of the image [1]. Several highly accurate image classification algorithms currently exist. This can be attributed to the availability of large databases of labeled images that are used to train and evaluate image classification algorithms. One such database is ImageNet [2], which includes over 14 million images. Each image in the ImageNet dataset is labeled with a WordNet synset representing the specific object in the image. A total of 21,841 unique synsets are used to label all of the images in the database. Figure 1 includes examples of images in ImageNet and their labels. In ImageNet, some of the synsets used are more high-level such as "animal" and "tree," while others are more specific such as "Maltese dog" and "Joshua tree." For a synset like "Maltese dog," ImageNet includes 1,287 images, one of which is displayed in Figure 1. The images in ImageNet were collected from various online sources.

Scholars from domains such as information retrieval and artificial intelligence have been investigating research challenges related to image processing and classification using ImageNet. During an annual competition, the Large-Scale Visual Recognition Challenge, participants compete and complete several tasks in image processing using ImageNet [3]. As a result of utilizing this database, there has been progress in relevant research areas such as classifying moving objects in videos, segmenting objects in images, and clustering search results. Most of the current image classification research is in English. The focus on the English language may be attributed to the lack of databases, like ImageNet, that contain images labelled in other languages. While some previous work has focused on developing methods to classify images in languages other than English, image classification in Arabic remains largely uninvestigated. This work aims to explore image classification in Arabic as well as to provide results and datasets useful for research in this area.

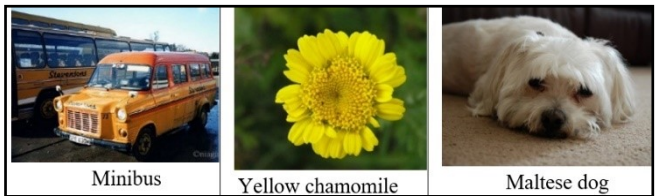
FIGURE 1. Sample of images in ImageNet and their labels.

The primary objectives of this work are: (1) to investigate the accuracy of using a common translation service to generate Arabic labels that are directly translated from the English labels or synsets available in ImageNet; (2) to provide initial baseline results for image classification algorithms for Arabic; and (3) to provide a dataset that consists of images labeled in Arabic based on the translation of labels in ImageNet. To achieve these objectives, a sample of 10,000 images is randomly selected from ImageNet. This dataset is reduced to 2,887 images after removing images that are no longer accessible online. An online English-to-Arabic translator is used to translate ImageNet's labels to Arabic. Finally, evaluation of the translation is conducted using a human judge to measure the accuracy of the generated Arabic labels for the images.

Exploring image classification in Arabic is important for several reasons. First, image classification algorithms can be incorporated into several applications for a wide range of technologies developed for individual users. For example, these algorithms can be used to help blind or visually impaired individuals identify physical objects present in the world. An Arabic speaking person may be unable to use an application that solely recognizes and speaks the names of objects in English. Another application is the real-time labeling of objects present in vehicles' dash board cameras. Such applications can be used by drivers as well as incorporated into road safety and monitoring systems. Non-English speakers will not be able to utilize such systems. Thus, image classification in Arabic could be beneficial in enhancing several applications that have the potential to improve accessibility and usability for Arabic speaking individuals. Furthermore, such techniques applied to Arabic could be extended to additional languages such as Chinese and Russian. Second, exploring image classification in Arabic is important because scholars in the domain of natural language processing (NLP) often develop solutions specific for target languages when they attempt to solve common NLP-related tasks. Examples of these tasks include document classification and summarization [4]–[7]. As each language has its own set of unique challenges, investigating solutions in a specified language could potentially highlight latent issues previously unnoticed. Therefore, targeting image classification in Arabic may uncover difficulties of image classification in general, and specifically applied to Arabic.

I hypothesize that state-of-the-art image classification algorithms that rely on a training dataset collected from a database such as ImageNet (with English labels) should perform similarly well when the underlying training dataset has images with labels written in other languages such as Arabic or Chinese. In other words, the original language of the labels in the dataset should not significantly affect the accuracy of the classification algorithm. To investigate this hypothesis, large datasets of images labeled in multiple languages would be needed. This work does not investigate this hypothesis. Instead, it provides a large dataset with images labeled in Arabic and English that can be used to examine this hypothesis and similar ones.

Alternatively, providing satisfactory results demonstrating the successful translation of image labels in ImageNet to Arabic may indicate that successful image classification may be achieved solely using translation services. This work seeks to measure the accuracy of generated labels when English labels or categories in ImageNet are directly translated to Arabic. A high accuracy would suggest that using image classification algorithms to produce labels in English and then translating the labels to Arabic could produce highly accurate results. In contrast, low accuracy results would suggest that alternative methods should be considered for successful image classification in Arabic.

This work makes several contributions. First, this is one of the only studies to focus on image classification in Arabic. Therefore, the results can be used as a baseline for future image classification in Arabic. Second, because the majority of image classification research is in English, this work adds to the limited research on image classification in other languages. Additionally, it is the first study to examine the accuracy of direct translations of ImageNet's labels to other languages. If the accuracy of direct translations of labels from English to Arabic is high, similar techniques can be applied to translate labels to additional languages. Finally, this study provides a dataset of 1,910,935 images with accurate Arabic labels for the objects in the images. Hence, this data can be used to test and evaluate other image classification in Arabic solutions.

The rest of the paper is organized as follows. Section II provides a review of the relevant literature; Section III details the methodology of this work; Section IV explains specific parameters of the experiment; Section V includes the results and discussion; Section VI presents the conclusions and future research.

## II. RELATED WORK

### A. IMAGE CLASSIFICATION
Image classification is the task of identifying and labelling an object or a set of objects present in an image. Recent advances in the field can partly be attributed to the availability of large-scale image datasets such as ImageNet [2]. ImageNet has accelerated the progress of artificial intelligence research on a broader scale, and more specifically image classification research [8]–[11]. Improvements in the performance of recent image classification methods are also attributed to the use of convolutional neural networks [12]. Many of these newer methods focus on "zero-shot" learning where objects are recognized even if they are not present as labeled data in the training dataset [13]. Other tasks related to image classification such as object detection and object tracking have also seen major advances. The goal of object detection is to find the boundaries of multiple objects in images. Highly accurate object detection algorithms include YOLO9000 [14] and R-FCN [15]. As for object tracking, the goal is to track the movement of an object in a scene, and recent work have been promising [16], [17].

Identifying objects present in an image could be beneficial in the development of several text-based information retrieval applications. For example, image classification can be used to generate image captions. These captions are comprised of full sentences that describe the contents of an image (rather than captions that only list the objects in the image) [18], [19]. Another application is visual question answering [20], [21]. The objective of this task is to answer a question about an image in a natural language. For example, when viewing an image of two teams playing soccer, a question could be "what are the colors of the soccer teams' shirts?" A successful answer would contain the correct colors. There are also several applications in specific industries. For example, in the healthcare industry, image classification methods can be used to generate medical text reports and relevant keywords based on image contents [22], tasks that are undoubtedly important. Although image classification research has important applications, the focus has traditionally been in English, not other languages. This limitation must be addressed because non-English speakers may directly interact with image classification applications. Therefore, future research should focus on studying aspects of current state-of-the-art methods in other languages.

### B. ARABIC NATURAL LANGUAGE PROCESSING

Arabic is one of the most commonly spoken languages around the world, thus, it is important to study applied computational solutions in the Arabic language. Scholars have studied various problems related to processing and analyzing texts in Arabic. Although several of the problems overlap with common NLP tasks, there are some issues specific to Arabic that have been previously investigated including developing methods for named entity recognition in Arabic [23], [24], sentiment analysis [25], [26], word segmentation [27], and question answering systems [28], [29].

Several scholars have discussed the difficulties associated with developing NLP methods and algorithms for Arabic as well as the need to build tools customized for the language [30]. These challenges include the ambiguity and complexity of Arabic[6], [31], the use of different Arabic dialects with unique characteristics [32], [33], and the limited number of freely available datasets that can be used in the research and development of Arabic information retrieval and processing solutions [34].

### C. IMAGE CLASSIFICATION FOR ARABIC

Image classification research has focused primarily on English with limited work on other Latin languages. One paper introduced a dataset of German words and their mapping to synsets in ImageNet titled "BilderNetle" [35]. To create the dataset, five native German speakers, one native English speaker, and a German-to-English translator were used to provide the German words. German translations were provided for 309 words and word-synset mapping to 2,022 synsets. ImageNet includes 21,841 synsets [36].

A limited number of related papers exist for Arabic. In one paper, crowdsourcing, a professional translator, and Google Translate were used to create a dataset that consists of 3,427 images and their Arabic captions [37]. This dataset was built to test a proposed image captioning model based on RNN-LSTM and CNN. Another paper also focused on generating full Arabic captions for images [38]. In this paper, a convolutional neural network was used to generate full sentences in Arabic that describe the contents of images. Arabic root words were incorporated in the training set. This method achieved a BLEU-1 score of 65.8 when it was tested on the Flicker8k dataset (with Arabic labels that were written by Arabic translators) and a BLEU-1 score of 55.6 when it was tested on 405,000 captioned images scraped from Arabic websites. BLEU is an evaluation metrics that is often utilized in machine translation and similar tasks [39]. The author indicated that this method performed better than when common image classification methods generated full captions in English, and a translation service then translated these captions to Arabic. While these results are promising, no details were provided regarding the translation or evaluation process.

### III. METHODOLOGY

In this section, the methodological details of this study are described. Figure 2 provides an overview of the major steps starting with ImageNet. The first step was to randomly select 10,000 images from ImageNet. Following that, the sample was reduced to 2,887 images using inclusion criteria determined before the study. Then, English labels for the images in ImageNet were translated to Arabic using an online translation service. The translated labels were evaluated by a human judge to determine if they accurately described the objects in Arabic. Details of how ImageNet was used, the translation process, and the specifications of the dataset as well as its limitations are in the following subsections.

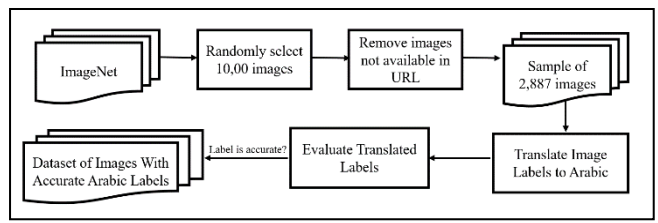

**FIGURE 2.** Overview of the methodological procedures for this study.

### A. IMAGENET AND DATASET

The dataset used in this study was constructed from the Fall 2011 release of ImageNet (multiple releases or versions of the datasets currently exist) [2]. ImageNet consists of 14,197,122 annotated images. In ImageNet, multiple objects can be present in a single image. For instance, a picture with a cat sitting on a table has at least two objects: a cat and a table. Furthermore, the database includes data on bounding boxes that define the locations of objects within a given image. In this paper, the bounding boxes and the existence of multiple objects in an image are not considered as they are outside of the scope of the study. Instead, only one category for each image as identified by ImageNet is used. To the best of my knowledge, ImageNet's release used in this study only includes one object for each image. In ImageNet, several attributes are available for each image. These attributes provide additional information on the given image. Prominent attributes include the synset in WordNet for the object in the image as well as the URL of the webpage where the image was initially downloaded. WordNet is a database that includes English words and their multiple definitions.

In WordNet, words may have multiple synsets where each is a unique definition or sense of a given word. Words with one definition only have one synset. Each synset includes the synset's part of speech such as a noun or a verb. For example, for the noun set of synsets for the word "chair," one definition is about the piece of furniture, while another refers to the academic position. Each synset also includes all of the synonyms for the specific word. For the second definition of chair, "professorship" is also included in the synset as it is a synonym for "chair" when used to refer to the academic position of a professor.

In ImageNet, WordNet's synset identifier is included with the identified object. Therefore, for an image of a person sitting on a chair, the ID of that synset is included to indicate that the synset of "chair" in WordNet that refers to the piece of furniture is used. In this study, the synset IDs were not incorporated in the translation process because the translation API used can only process words. There is no clear method to provide synsets' information for the translation service to process.

Arabic WordNet is a similar resource to English WordNet. Words and synsets in Arabic WordNet are mapped to the matching words and synsets in English WordNet. To test whether Arabic WordNet can be used to translate labels in ImageNet, an experiment was conducted using a sample dataset. The details of this experiment are in Section IV.

To build the dataset used in this study, a randomly selected sample of 2,887 images in ImageNet was used. To construct this sample, a larger sample of 10,000 images was randomly selected from ImageNet. Because the image URLs were used to view the images and display them in an online platform in which the judge labeled the results, a python script was used to determine if each image was accessible online and therefore useable for this study. In ImageNet, the images were collected from various websites including image hosting websites such as Flickr. After removing images that were no longer accessible online, the sample of images was reduced to 2,887 images. The high percentage of images in the sample of 10,000 images that were no longer available online (71.1%) suggests that the URL links of images in ImageNet are not reliable when used, for example, to display the entire dataset. It was observed that links from the website Flickr are more likely to remain available online when compared to other hosting websites. However, no detailed investigation was conducted to examine the accuracy of this statement.

### B. TRANSLATION PROCESS

For the translation process, the API of Google Translate was used. Google Translate is an online service where an input text written in a particular language can be translated to another selected language. No information was found on the accuracy of the API when used to translate text from English to Arabic. However, according to one study, when compared to three other online translation services including Microsoft Bing, Google Translate was more accurate when translating sentences from English to Arabic [40]. For this reason and because of its reliability and (perceived) popularity, Google Translate was used in this study. Using other online translation services or human translators to translate the labels of images in ImageNet from English to Arabic may produce results that are different from the ones found in this study. This translation API was used to translate all of the labels for the sample of 2,887 images to Arabic.

### C. NEW DATASET OF IMAGES WITH ARABIC LABELS

Following the evaluation process (which is explained in the experiment section of this paper), all of the images with a correct Arabic translation were added to a separate dataset comprised of only images with correct Arabic labels. This dataset represents a novel contribution of this paper. Because each of the 21,841 synsets in ImageNet is linked to many images, the small dataset generated in this study was extended by including all of the additional images of a synset with correct translations.

There are several limitations of this dataset. First, it can be argued that the dataset only contains images that were perhaps easier to translate from English to Arabic. Second, the generated dataset may include a smaller percentage of fine-grained categories than ImageNet. Third, one of the advantages of ImageNet is that it uses WordNet's hierarchy of words and synsets to label images. This is accomplished by labeling images with the appropriate WordNet synsets that best describe the objects in the images. The dataset generated in this study includes only words in Arabic with no direct links that maintain WordNet's semantic structure. Nevertheless, while these limitations must be considered, this dataset is a good resource that can be utilized in future studies in the domain. This dataset and the dataset of images with incorrect labels are both available upon request.

## IV. EXPERIMENT

In this section, the details of using Arabic WordNet to translate labels in ImageNet and a description of the evaluation process are explained.

### A. ARABIC WORDNET TESTING

Arabic WordNet is a lexical resource similar to English WordNet [41]. However, it is still considered a limited resource and few scholars have attempted to enhance its contents [42]. One feature of Arabic WordNet is the mapping between synsets in Arabic WordNet to synsets in other languages, including English. This raises the following question: If there is direct mapping between synsets in English WordNet to those in Arabic WordNet, can Arabic WordNet be used to translate all of the synsets in ImageNet?

To answer this question, a sample of 100 images was randomly selected from ImageNet. For each image, an attempt was made to find the Arabic synset linked to the image's label in ImageNet. Arabic WordNet was used for this search. For the 100 images, a corresponding Arabic label was only found for six images. Thus, while Arabic WordNet should be further explored as a viable tool for image classification in Arabic, it was demonstrated that it can only be used to translate a limited number of images from ImageNet.

### B. TRANSLATIONS EVALUATION PROCESS

Evaluation of the performance of the translation service was completed by a human judge who is fluent in English and Arabic. The judge was responsible for evaluating the accuracy of each translated label by indicating if a translation was "accurate," "inaccurate," "neutral," or provided in English rather than Arabic. The judge used several external resources that provided additional information about the translations.

Several Arabic dictionaries were used when the judge was examining the accuracy of the translations. The judge referenced the dictionaries in the two following scenarios. First, one of the characteristics of ImageNet is the fine-grained aspect or specificity of some of the categories generated for objects. For example, there are images labeled with the synsets "great blue heron" or "trogon" in addition to images labeled with the high-level category "bird." The Arabic translations for these fine-grained labels may be unknown to the judge. Second, the translation service produced results in Modern Standard Arabic (MSA), and the definitions of these results may not be widely known. In these scenarios, Arabic dictionaries were used to provide definitions for unclear translations and further clarification. The need to use a dictionary made the evaluation process labor intensive.

Another external resource used by the judge was an online version of WordNet [43]. English WordNet was used to obtain full definitions of the synsets used in ImageNet. In addition to Arabic dictionaries and WordNet, the judge used other resources to aid in the evaluation process. These resources were used to gather additional information on the synsets, their translations, and the full definitions for translations.

The judge completed the evaluation process using an online datasheet. The datasheet included the following information: the unique identifiers for the images, the synsets (labels) for the images in English as appeared in ImageNet, the labels for the image in Arabic as translated by the translation service, and the embedded images that can be viewed in the sheet. The judge also noted whether the translation was "accurate," "inaccurate," "neutral," or provided in English rather than Arabic.

Labels were categorized as "accurate" when the generated Arabic labels correctly described the objects in the images. Labels were categorized as "inaccurate" when the Arabic labels did not accurately describe the objects in the images. Labels were categorized as "neutral" when there was insufficient information to confidently decide whether a translation was "accurate" or "inaccurate." Finally, the label of "English" was used when the translation API's output provided a word in English, not Arabic. For example, the API translated the word "barouche" to "barouche." Although it is unclear why this occurs, a manual inspection of the data suggests that there are not equivalent Arabic words for these English words. For some translations, the text included both an Arabic word and an English word. These instances were also labeled with the category "English."

To measure the accuracy of using a translation service to translate labels in ImageNet from English to Arabic, the total number of accurately and inaccurately generated labels in Arabic were used. Images that were labeled as "inaccurate," "neutral," or "English" were grouped together as "inaccurate" whereas images that were labeled correctly were classified as "accurate" labels. To quantify the performance, the number of "accurate" instances was divided by the total number of images in the dataset.

## V. RESULTS AND DISCUSSION

This section describes the results of using a translation service to translate a sample of images in ImageNet from English to Arabic, the datasets generated in this study, the summary of the effects of the textual structures of the labels on translations. Additionally, observations about the results and how they can be improved are provided by highlighting several observed categories of outputs.

### A. TRANSLATIONS RESULTS

TABLE 1
SUMMARY OF RESULTS FOR THE GENERATED LABELS

| Result | Number of Images | Percentage |
|---|---|---|
| Accurate | 1,895 | 65.63% |
| Inaccurate | 728 | 25.11% |
| Neutral | 74 | 2.56% |
| English | 190 | 6.58% |

| | 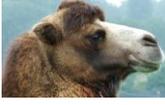 | 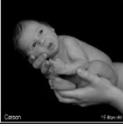 | 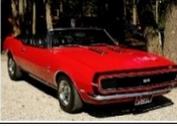 |
|---|---|---|---|
| ImageNet label | Camel | Neonate | Sports car |
| Our label | جمل | حديثي الولادة | سيارة رياضية |

**FIGURE 3.** Sample of Images with Correct Translations.

Based on the judge's evaluation, the results indicated that translated Arabic labels for 1,895 of the 2,887 images were "accurate." Therefore, 65.6% of the labels were translated successfully. This percentage represents the number of generated Arabic labels that accurately captured the objects in the images. Alternatively, 34.4% of the translated labels did not accurately describe the objects in the images. This included labels classified by the judge as "inaccurate," "neutral," and written in English. Table 1 includes a summary of the results and the number of images in each of the four possible classes of labels.

The judge classified 74 of the translations as "neutral" because they could not definitively indicate whether the generated label was "accurate" or not. For example, for an image of the plant "jonquil," the judge was unable to confidently determine if the generated Arabic label was "accurate." Other examples of neutral labels include images of the plant "hepatica" and the insect "atlas moth." Additionally, the Arabic words were uncommon for several of the "neutral" labels and the judge was not able to find definitions for them.

In Figure 3, three images with correct translations for the labels are displayed. The "ImageNet labels" are the synsets present in ImageNet. "Our labels" are the labels generated by the translation service. For example, the first image in Figure 3 is of a camel. The Arabic translation is "جمل" which is the Arabic word for camel.

### B. DATASET RESULTS

Based on the completed evaluation, 1,895 images from ImageNet with "accurate" Arabic labels were identified. These images were labeled with 1,643 unique synsets. In ImageNet, each of these synsets is linked to many images. For example, the synset "coffee.n.01" is linked to 1,283 different images. In ImageNet, each synset is linked to an average of 1158.1 images. To extend the dataset of images from ImageNet with "accurate" Arabic labels, the 1,643 successfully translated synsets were linked to the same images that each synset is linked to in ImageNet.

This extension expanded the dataset from 1,895 images to a new dataset with 1,910,935 images (with 1,643 unique labels). The new dataset can be used in future research involving tasks related to image classification in Arabic. Each image in the new dataset is associated with one of the 1,643 successfully translated categories.

TABLE 2
SUMMARY OF RESULTS BY SYNSETS AND IMAGES

| Type | Accurate Arabic Labels Generated | ImageNet | Percentage |
|---|---|---|---|
| Synsets | 1,643 | 21,841 | 7.5% |
| Images | 1,910,935 | 14,197,122 | 13.4% |

Table 2 provides summary of results of this extension of the dataset. Since ImageNet is comprised of a total of 14,197,122 images, the new dataset provides valid Arabic translations for 13.4% of the images available in ImageNet. Furthermore, an Arabic translation is available for 7.5% of the synsets in ImageNet.

### C. TEXTUAL STRUCTURE OF LABELS

To investigate whether the textual structure of the labels in ImageNet has various effects on the performance of using a translation service to translate image labels in ImageNet to Arabic, the dataset was divided into smaller sections based on the number of words in the label's name. Three classes were created: "unigrams," "bigrams," and "n-grams." The first class of "unigrams" included labels with only one word; the second class of "bigrams" included labels with two words; the third class of "n-grams" included labels with three or more words. The dataset was divided into these three categories to identify possible differences in performance for each class. It is possible that the accuracy of the translation would be higher for labels consisting of only one word.

Results varied based on the three types of textual structures of labels (unigrams, bigrams, and n-grams). The performance was highest when the labels were "unigrams." In these instances, 71% of the translated labels were classified as "accurate" by the judge. Percentages scores of successfully translated labels for "bigrams" and "n-grams" were lower, 58% and 45% respectively. Figure 4 provides a summary of the performance for "unigrams," "bigrams," and "n-grams." As the number of words increased, the number of "accurate" labels decreased, and "inaccurate" labels increased. However, for "n-grams," the percentages of "accurate" and "inaccurate" labels were similar.

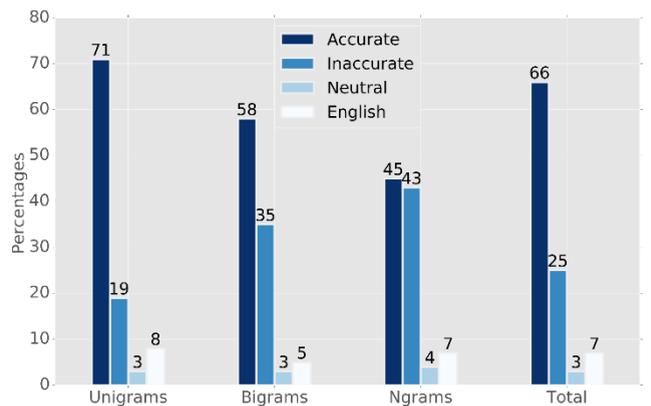

**FIGURE 4.** Performance for "unigrams," "bigrams," "n-grams," and the three categories combined. The performance decreased when the labels consisted of multiple words.

TABLE 3
SUMMARY OF RESULTS FOR UNIGRAMS, BIGRAMS, AND N-GRAMS

| Category | Number of Images | Unique Synsets | Percentage of Accurate Translations |
|---|---|---|---|
| Unigrams | 1,288 | 1,095 | 71% |
| Bigrams | 576 | 519 | 58% |
| N-grams | 32 | 29 | 45% |

Table 3 shows the results for the three categories ("unigrams," "bigrams," and "n-grams") based on the overall number of images in each category. For the 1,895 accurately translated labels, 1,288 were "unigrams" with 1,095 unique synsets, 576 were "bigrams" with 519 unique synsets, and 32 were "n-grams" with 29 unique synsets.

One factor that could have attributed to the decrease in performance for "bigrams" and "n-grams" is that several of the English "bigrams" and "n-grams" were translated to Arabic as a set of unrelated words instead of a single unit or a noun phrase. For example, for phrases such as "reading teacher," the Arabic translation provided separate translations for "reading" and "teacher." The translations of both words were combined in the full translation as if the two words were unrelated. Another similar phrase was "hurricane deck." This is one possible explanation for the decrease in the number of accurate translations for "bigrams" and "n-grams." However, future studies should be conducted to determine if additional factors contribute to the lower percentage of "accurate" translations for "bigrams" and "n-grams." New solutions to provide Arabic labels for images in ImageNet should attempt to create methods to eliminate such issues.

### D. OBSERVED CATEGORIES

In addition to the translations that were clearly correct, four other types of translations were observed during the evaluation process. These four additional categories reveal common errors in the translation process as well as areas where novel solutions related to image classification and information retrieval can be explored. Several examples of translations found in these observed categories are displayed in Figures 5, 6, 7, and 8.

The presence of these categories suggests that minor modifications and preprocessing of English labels prior to the use of a translation service may increase the overall performance of using a translation service to generate Arabic labels for images in ImageNet. Although the four categories represent different types of observed categories, there may be additional categories that were not included in this section. One additional category that is not included in this section was a set of labels translated to Arabic that also included one or more English words. Another unexplored category included noun phrases in English that were translated as a group of unrelated words in Arabic. The following four subsections include descriptions of the four, explored categories as well as several examples.

### 1) INCORRECT SYNSET

Word sense disambiguation refers to the task of identifying the right sense used in text for a word. English words commonly have several senses or definitions that refer to the different uses for a given word. In WordNet, each sense of a word is linked to a synset that also includes all the synonyms of that sense. ImageNet provides the synset's identifier in WordNet for each category or label. For example, for the image of a skunk in Figure 5, ImageNet specifies that the fourth noun synset for "skunk" in WordNet is used. The fourth synset's definition for skunk is an "American musteline mammal typically ejecting an intensely malodorous fluid when startled; in some classifications put in a separate subfamily Mephitinae."

In this paper, synset identifiers were not incorporated in the translation process as there is no direct method to provide such identifiers to the translation service. Implementing preprocessing steps that clarify the synset used by ImageNet prior to using a translation service may reduce the number of inaccurately translated labels. However, it is important to note that many image classification algorithms generate labels without providing the synset identifier in WordNet. Therefore, building a method for image classification in Arabic that depends on using the synset's identifier in WordNet will fail if the underlying image classification algorithms in English do not specify synset identifiers of the images in the results.

The first observed category of results includes images where the translation API translated the wrong synset of a given word. Results in this category were "inaccurate" because the translation API translated a different synset of the label or word. For example, for the image of a "skunk," the animal is present in the image. However, the API assumed that the word "skunk" was used to refer to obnoxious or unfriendly individuals who are described as "skunks." In such cases, the translations were labeled as "inaccurate." If a word has multiple synsets, it is unclear how the translation API determines which synsets to use. Providing specific information to ensure that the correct synsets of a word is used by the translation service may produce more accurate results. Furthermore, using a translation algorithm where the identified synset can be used as an input may improve the results.

| | 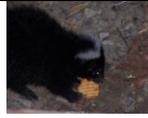 | 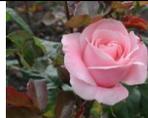 | 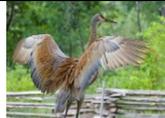 |
|---|---|---|---|
| ImageNet label | Skunk | Rose | Crane |
| Our label | شخص بغيض | ارتفع | مرفاع |
| English description of our label | Obnoxious guy | Rose as in rose in the air- past tense of rise | A machine used to move heavy objects |

**FIGURE 5.** Sample of images with incorrect synset.

### 2) FULL DEFINITIONS

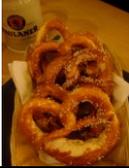

| ImageNet label | Pretzel | Croquette | Airedale |
|---|---|---|---|
| Our label | كعك مملح و جاف | الكروكيت كتلة من لحم سمك | الأردبل نوع من كلاب الصيد الضخمة |
| English description of our label | Dry and salty cake | Croquette a block of fish meat | Airedale a type of huge hunting dogs |

**FIGURE 6.** Sample of images with full definitions as translations.

The second observed category consists of images where the translation API provided a full definition in Arabic for the English label in ImageNet. In other words, the API translated a single word in English to a full sentence in Arabic providing a complete Arabic definition of the English word. For example, the translation for "pretzel" in Arabic is " كعك مملح و جاف" which can be translated back to English as a "dry and salty cake" rather than the Arabic word for "pretzel." Figure 6 provides additional examples of images in this category. It is unclear as to why full definitions were provided. However, it seems that such instances happen for English words that do not have directly matching Arabic words. In the evaluation process, images in this category were classified as "accurate" if the Arabic definitions were deemed correct by the judge.

### 3) CORRECT BUT UNCOMMON

The third observed category includes Arabic labels translated by the API that accurately describe the images. However, the words in this category are uncommon and rarely used in Arabic resources or by Arabic speaking individuals. These translations were classified as "accurate" only after the judge used an Arabic dictionary to identify the correct meaning of the words provided by the translation API. While the translations were accurate, these words may not be recognized by the average, native Arabic speaker. Thus, computational solutions that incorporate these translations may be problematic for the general Arabic speaking population. For many images in this category, it can be argued that the English words (WordNet synsets) as used by ImageNet are similarly uncommon and rare. Words such "earthwork" and "teasel," which are displayed in Figure 7, may be unknown to native English speakers as well.

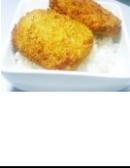

| ImageNet Label | Shutter | Earthwork | Teasel |
|---|---|---|---|
| Our label | مصراع | متراس | ممشقة |

**FIGURE 7.** Sample of images with "accurate" but uncommon words are used in the translation.

Several images in this category are part of the "fine-grained" category in ImageNet. All of the images in this category were classified as "accurate." Additionally, it is possible that several of the images that were classified as "neutral" in the evaluation process contain correct, but uncommon Arabic words that were unknown to the judge and could not be identified using an Arabic dictionary.

### 4) SAME WORD, DIFFERENT ALPHABET.

The fourth observed category includes images where the English labels were translated into the same, exact word, however spelled with letters of the Arabic alphabet. For example, for an image of "Tiramisu," the translation was the word "Tiramisu," but in Arabic letters rather than letters of the English alphabet (Figure 8). This presumably happens because some English words are added to the Arabic dictionary similarly to how some Arabic words such as "hummus" and "falafel" are entered into the English dictionary. It seems that the majority of the images in this category were names of specific food items. All of the images in this category were classified as "accurate" in the evaluation process.

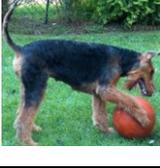

| ImageNet Label | Tiramisu | Sicilian Pizza | Cappuccino |
|---|---|---|---|
| Our label | تيراميسو | بيتزا سسيليا | كابتشينو |

**FIGURE 8.** Sample of images with the same words written in Arabic as a translation.

## VI. CONCLUSION

In this paper, an experiment was conducted to examine the accuracy of using a translation service to translate English labels in ImageNet to Arabic. A sample of 10,000 images were randomly selected from ImageNet. This dataset was reduced to 2,887 images after images that were no longer accessible were removed. The major finding of this study was that 65.6% of the images resulted in accurate translations. This finding can be used as a baseline accuracy level for other image classification methods in Arabic. Furthermore, this study provides a dataset of 1,910,935 images that are labeled with correct Arabic labels. All of the images belong to one of 1,643 unique labels or synsets. This dataset can be used in subsequent studies that target image classification in Arabic.

With additional modifications and the inclusion of preprocessing steps, using online translation services to translate labels of images to Arabic could produce better results. One common problem that occurred was when the incorrect synset of a word was used in the translation. Therefore, by providing contextual information about the images prior to translation, the accuracy of the translated labels could be higher. Another possible modification is one that focuses on fine-grained categories present in ImageNet. It was observed that such categories increase the percentage of incorrect translations. Examples of fine-grained categories include specific types of birds or dogs. Some of these animal breeds may not exist in Arabic speaking countries, and it is unknown if Arabic names for these categories exist. Considering alternative methods to translate fine-grained categories should increase the overall performance. One possible solution is to start by translating high-level synsets such as "bird" and "animal" and then examine their subcategories.

There are several directions for future research in this area. Research should primarily focus on how to provide Arabic labels for all of the images in ImageNet. To accomplish this objective, one possible approach is to translate all of the synsets to Arabic with a translation service and then use the evaluation method followed in this paper. Another possibility is to introduce modifications to the labels prior to or after the translation process with the inclusion of a preprocessing step. Furthermore, the inclusion of additional translation services could also increase the number of true positives. Future studies should investigate these directions.